\documentclass{styles/svproc}

\usepackage{url}
\usepackage{hyperref}
\usepackage[dvipsnames]{xcolor}
\usepackage[figuresright]{rotating}
\usepackage{amsmath}
\allowdisplaybreaks

\begin{document}
\mainmatter              
\title{A Modular Ontology for MODS -- Metadata Object Description Schema}
\titlerunning{Modular MODS Ontology}
\author{Rushrukh Rayan\inst{1} \and Cogan Shimizu\inst{2} \and Heidi Sieverding\inst{3} \and Pascal Hitzler\inst{1}}
\authorrunning{Rayan, Shimizu, Sieverding, Hitzler} 
%
%
\institute{Kansas State University, $\{\text{rushrukh,hitzler}\}$@ksu.edu \and Wright State University, cogan.shimizu@wright.edu \and South Dakota School of Mines \& Technology, heidi.sieverding@sdsmt.edu}

\maketitle  

\begin{abstract}

The Metadata Object Description Schema (MODS) was developed to describe bibliographic concepts and metadata and is maintained by the Library of Congress. Its authoritative version is given as an XML schema based on an XML mindset which means that it has significant limitations for use in a knowledge graphs context. We have therefore developed the Modular MODS Ontology (MMODS-O) which incorporates all elements and attributes of the MODS XML schema. In designing the ontology, we adopt the recent Modular Ontology Design Methodology (MOMo) with the intention to strike a balance between modularity and quality ontology design on the one hand, and conservative backward compatibility with MODS on the other. 

\end{abstract}



\section{Introduction}

XML -- a markup language -- is designed to organize information \cite{xml-tr}. The main design goal is to store and share information while maintaining human and machine readability. Also, the purpose of XML Schema is to serve as a description for an XML document, within it detailing the constraints on the structure, syntax, and content type. The schema outlines rules and constraints for elements, attributes, data types and relationships between them. It also helps ensure that the XML document conforms with the expected structure, serving as a way of validation. It is important to note that XML structures information in a hierarchical form, essentially representing a tree structure. \bigskip

The Metadata Object Description Schema (MODS) \cite{modsmetadata} is an XML schema developed by the Library of Congress' Network Development in 2002 to be used to describe a set of bibliographic elements. MODS contains a wide range of elements and attributes using which a well-rounded description can be provided about bibliographic elements. For instance, it has \textit{elements} to describe \textsf{Title Information}, \textsf{Type of Resource}, \textsf{Genre of Resource}, \textsf{Origin Information}, \textsf{Target Audience}, \textsf{Access Restrictions} of the material, etc. Furthermore, MODS also has \textit{attributes} to outline additional important information, to name a few: \textsf{Display Label} (describes how the resource under description is being displayed), \textsf{Lang} (points to the language that is used for the content of an element: imagine a book title that is French), \textsf{Authority} (specifies the organization that has established the usage of, for instance, an acronym), etc. General example use-cases of MODS lie within the realm of describing metadata of Journal Publications (one or more), Research Projects, Experiments, Books, etc. \bigskip

While XML schema does a decent job in imposing structure on XML data, it lacks some desirable features. In the age of data, where cleaning, pre-processing, and managing data takes up a large chunk of resources in data operation, it is desirable to have the ability to organize data in such a way that allows semantic expressiveness of the data and conveys information on relationships between various \textit{concepts} by means of a \emph{graph structure} \cite{Hitzler2010} as opposed to the XML tree structure, in the sense of modern knowledge graphs \cite{Hitzler21}, e.g. based on RDF \cite{rdf-tr} and OWL \cite{owl-tr}. 
An XML schema
\begin{itemize}
    \item lacks semantic expressiveness to convey relationship among concepts, context of data;
    \item lacks native support for automated reasoning and inference;
    \item lacks a common framework that allows integration of data from various sources;
    \item possesses a hierarchical nature with a rigid structure which makes it rather less flexible with respect to incorporation of different perspectives;
    \item and lacks native support for querying.
\end{itemize}
Ontologies as knowledge graph schemas, on the other hand, provide a structured and graph-based way to represent knowledge in an application domain. By defining the necessary vocabulary, concepts, entities, and relationship between concepts, ontologies allow a meaningful interpretation of the data. 

The reason we have developed the Modular MODS Ontology (MMODS-O) is to address some of the challenges which the MODS XML schema exhibits. Indeed MMODS-O is designed to strike a balance between conservative backward-compatibility with the MODS XML schema and quality modular ontology design principles following the MOMo methodology \cite{momo-swj,modl}. 
The modular structure in particular is supportive of simplified extending, modifying or removing parts of the ontology.

We have created 34 modules and patterns to capture the entire MODS XML schema. To provide semantic robustness, we have re-engineered some of the modules from their XML schema definition. The schema is expressed in the form of an OWL Ontology and extensive documentation is available on Github\footnote{\url{https://github.com/rushrukh/mods_metadata_schema/tree/main/documentation}}. 

One of our target use-cases for MMODS-O is to provide a metadata structure to a large-scale collaborative research project, where the knowledge graph would contain information such as different research groups, experiments performed, geo-location information, associated publications, presentations, book-chapters, collaborators etc.

We would like to point out that this is not the first attempt towards developing a MODS Ontology. However, our version is an improvement over the existing ontology across multiple aspects, including modular structure, adherence to MOMo quality control principles, rich axiomatization, extensive documentation. We will outline some of the key improvements over previous work in Section \ref{sec:comparison}. In general, our contributions are: 

\begin{enumerate}
    \item Development of the modular ontology, where some of the modules differ significantly from the original MODS XML schema in order to reflect good ontology design principles.
    \item Carefully considered and rich axiomatization to scope intended usage and to provide automated reasoning capabilities.
    \item Complete documentation of the graph schema outlining each of the modules, associated axioms, competency questions.
\end{enumerate}

The rest of the paper is organized as follows. Section \ref{sec:modules} contains the description of key modules from our ontology. 
In Section \ref{sec:comparison}, we describe related work and highlight some of the key differences of our modeling with previous efforts. 
We conclude in Section \ref{sec:conc}.
The ontology is available as serialized in the Web Ontology Language OWL from \url{https://github.com/rushrukh/mods_metadata_schema/tree/main/modules}.



\section{Description of the MODS Ontology}
\label{sec:modules}

The general usage of the MMODS-O (and MODS) lies in the realm of expressing bibliographic metadata. Indeed, the details in the XML schema reflect the association with bibliographic applications. From the top level elements and their attributes in the MODS XML schema, we have identified 34 modules to be part of MMODS-O. Some of the key modules are briefly described below. The primary goal of using formal axiomatization\footnote{A primer on description logic and the notation can be found in \cite{dlprimer,Hitzler2010}} in MOMo is to limit unintended use and to disambiguate the modules, but axioms can also be used for logical inferences \cite{DBLP:books/ios/p/HitzlerK16}. The axioms are expressed using the OWL 2 DL profile \cite{owl-tr}. Note that for all the modules outlined here, the list of axioms is not complete as we only highlight some of the most important axioms for brevity. The complete list of axioms and modules can be found in the documentation pointed to earlier.

The modules that we selected for presentation in this paper include some that deviate most from the underlying MODS XML schema. We touch upon the differences throughout and will discuss them further in Section \ref{sec:comparison}.

We make extensive use of schema diagrams when discussing modules following the suggested visual coding from the MOMo methodology \cite{momo-swj} where further explanations can be found: orange (rectangular) boxes indicated classes; teal (dashed) boxes indicate other modules (and usually also the core class of that module); purple (dashed) boxes with \emph{.txt} indicate controlled vocabularies (i.e., formally, classes with pre-defined individuals as members, which have meaning that is defined outside the ontology); yellow (ovals) indicate datatype values; white-headed arrows are rdfs:subClassOf relationships, all other arrows are object or data properties, depending on type of node pointed to.

\subsection{Overview of the Modules in the Ontology}
\begin{figure}[h!]
    \centering
    \includegraphics[width=\textwidth]{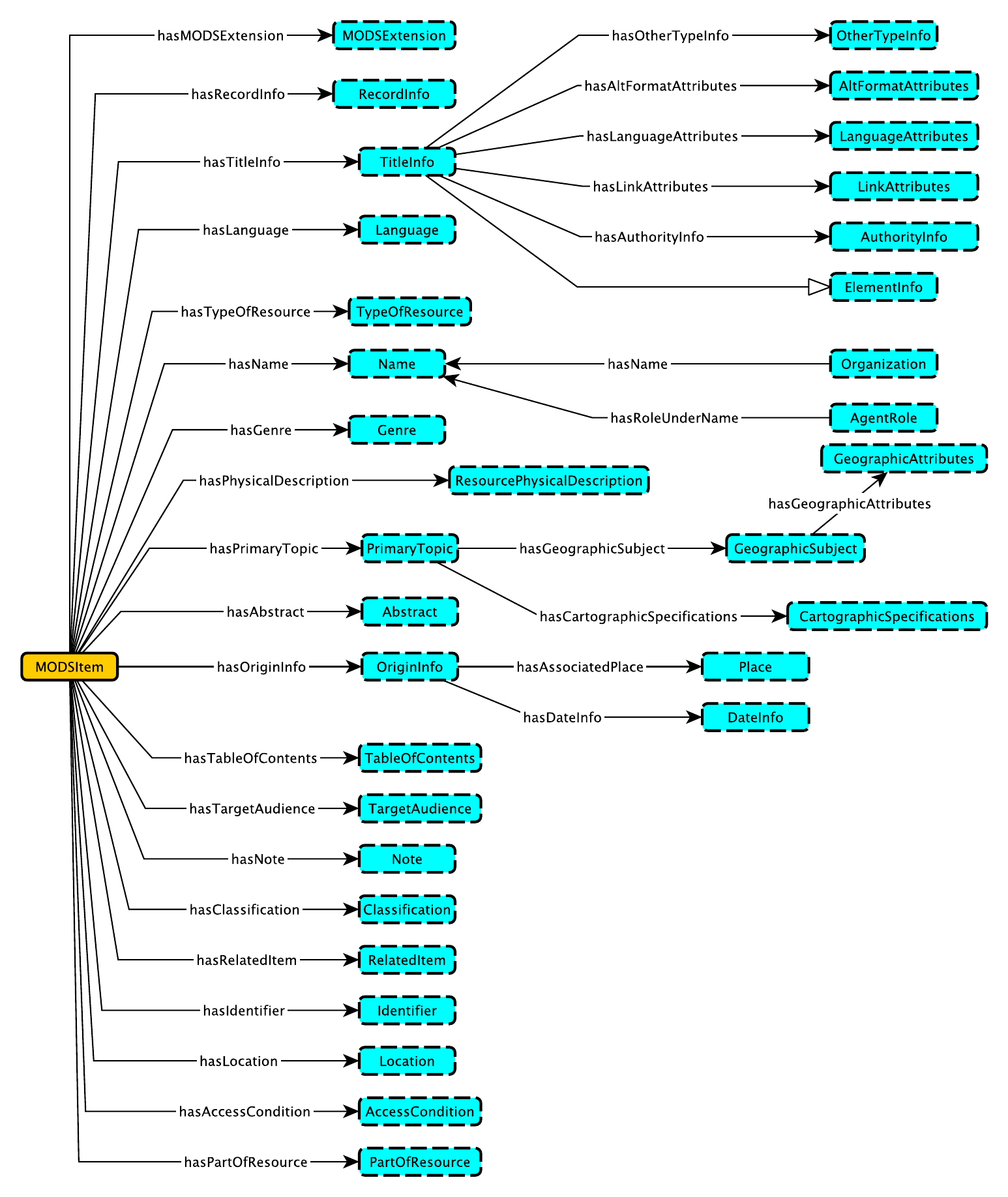}
    \caption{An Overview of all the Modules}
    \label{fig:overview-modules}
\end{figure}

Figure \ref{fig:overview-modules} represents a brief overview of all the modules that are part of the ontology. Each of the modules has its separate schema. MODS Item is a reference to the MODS resource under description. The ontology has 34 modules, while we highlight some of the key modules later in the paper, details about the other modules are available in the documentation. Figure \ref{fig:overview-modules} suggests almost a tree structure, which is actually not the case but this is not quite apparent from this high-level perspective.

\subsection{Role-Dependent Names}

Role-Dependent Names is an ontology design pattern \cite{DBLP:books/ios/p/HitzlerK16,momo-swj} that is useful when there is an \textsf{Agent Role} that is performed by an \textsf{Agent}. Naturally, Agent will have a \textsf{Name}. There are instances when an Agent assumes a Role under a particular Name, but the same Agent will assume a different role under a different Name. An example for such a scenario would be a writer writing different books under different pseudonyms. For example, Ian Banks publishes science fiction as "Iain M. Banks" and mainstream fiction as "Iain Banks".  Another example use case within the application scope we are primarily interested in could be as follows: if the resource under description refers to a journal publication, there would be Agent Roles for authors, which would be assumed by Agents under some name. Note that names associated with an author may differ between different publications for a variety of reasons, including different transcriptions from other languages, inclusion or not of middle names, name changes, etc., and the MODS XML schema reflects this. While we do not discuss the ontology design pattern at length here, details can be found in \cite{roledependentname}.

\begin{figure}[h]
    \centering
    \includegraphics[width=\textwidth]{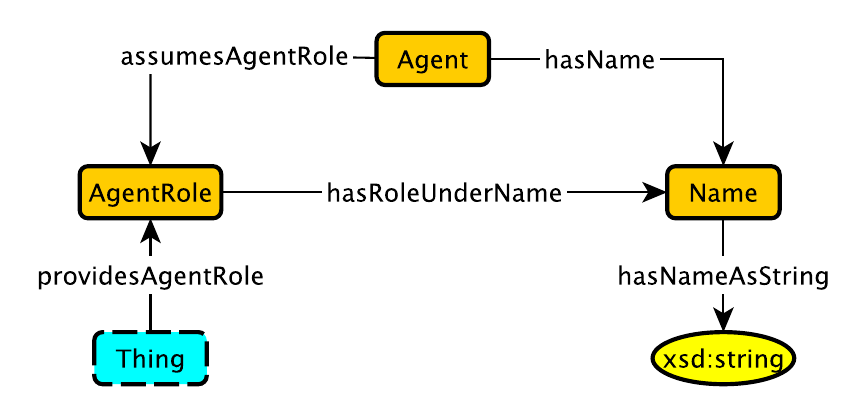}
    \caption{Schema Diagram for the Role-Dependent Names Pattern}
    \label{fig:agentrolename}
\end{figure}

\subsubsection{Selected Axioms}
\begin{align}
\top &\sqsubseteq \mathord{\leq} 1\textsf{providesAgentRole}^-\textsf{.} \top \label{agentrole1} \\
\textsf{AgentRole} &\sqsubseteq \mathord{\geq} 0\textsf{hasRoleUnderName.Name} \label{agentrole2}\\
\exists \textsf{assumesAgentRole.Agent} &\sqsubseteq \textsf{AgentRole} \label{agentrole3}\\
\textsf{AgentRole} &\sqsubseteq \mathord{\leq} 1\textsf{assumesAgentRole}^-\textsf{.Agent} \label{agentrole4}\\
\textsf{Agent} &\sqsubseteq \mathord{\geq} 0\textsf{assumesAgentRole.AgentRole} \label{agentrole5}\\
\textsf{Agent} &\sqsubseteq \exists \textsf{hasName.Name} \label{agentrole6}\\
\textsf{assumesAgentRole} \circ \textsf{hasRoleUnderName} &\sqsubseteq \textsf{hasName} \label{agentrole7}\\
\textsf{hasName} \circ \textsf{hasRoleUnderName}^- &\sqsubseteq \textsf{assumesAgentRole} \label{agentrole8}
\end{align}

If an Agent Role is provided, we argue that there must be at most~1 entity that provides the role which is expressed using an \textit{inverse functionality} in (\ref{agentrole1}). Furthermore, we claim that if an Agent Role is assumed, there can be at most~1 Agent who assumes the role, expressed through an \textit{inverse qualified scoped functionality} in (\ref{agentrole4}). Axioms (\ref{agentrole1}) and (\ref{agentrole4}) essentially state that an AgentRole is unique to both the Agent and the entity providing the role, i.e., these axioms give guidance as to the graph structure for the underlying data graph. It is not necessary for an Agent Role to be assumed under a Name which is why we use a \textit{structural tautology} in (\ref{agentrole2}).\footnote{Structural tautologies are logically inert, however they provide structural guidance on use for the human using an ontology; see \cite{momo-swj}.}  We also argue that, naturally, an Agent must have a name. Hence we use an \textit{existential} to convey that in (\ref{agentrole6}).

The Role-Dependent Names module exemplifies very well why an RDF graph structure is much more natural than an XML tree structure for expressing relevant relationships. In particular, the \emph{triangular} relationships indicated by the role chain axioms (\ref{agentrole7}) and (\ref{agentrole8}) cannot be naturally captured in a tree structure, but really demand a directed graph.

\subsection{Element Information}
There are many elements within the MODS XML schema which may have a display label, a combination of attributes that provide external links, and a set of attributes to describe the language for the resource under description.
The Element Information module is created such that the aforementioned connections can be expressed conveniently. Concretely, whenever in a module it needs to be said that the module may have a Display Label, Link Attributes, and Language Attributes, we use the module to be a sub-class of the module Element Information which is expressed using a \textit{sub-class of} relationship in (9).

\begin{figure}[h]
    \centering
    \includegraphics[width=\textwidth]{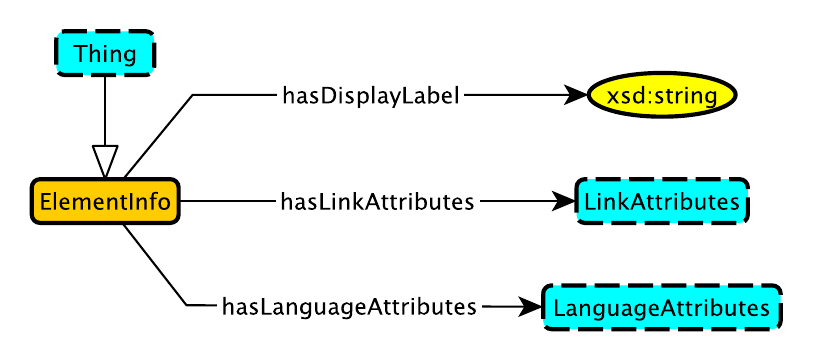}
    \caption{Schema Diagram for the Element Information Module}
    \label{fig:elementinfo}
\end{figure}

\subsubsection{Selected Axioms}
\begin{align}
  \top &\sqsubseteq \textsf{ElementInfo} \label{elementinfo1}\\
  \top &\sqsubseteq \mathord{\leq} 1\textsf{hasLinkAttributes.} \top \label{elementinfo2}\\
  \textsf{ElementInfo} &\sqsubseteq \mathord{\geq} 0\textsf{hasLinkAttributes.LinkAttributes} \label{elementinfo3}\\
  \top &\sqsubseteq \forall \textsf{hasLanguageAttributes.LanguageAttributes} \label{elementinfo4}\\
  \top &\sqsubseteq \mathord{\leq} 1\textsf{hasLanguageAttributes.} \top \label{elementinfo5}\\
  \textsf{ElementInfo} &\sqsubseteq \mathord{\geq} 0\textsf{hasLanguageAttributes.LanguageAttributes} \label{elementinfo6}
\end{align}

A module which is a sub-class of Element Information can have at most 1 set of Link Attributes and 1 set of Language Attributes which in axioms have been conveyed using \textit{functionalities} in (\ref{elementinfo2}) and (\ref{elementinfo5}). Additionally, it is not mandatory for a module to have a set of Link Attributes and Language Attributes, therefore we make use of \textit{structural tautologies} in (\ref{elementinfo3}) and (\ref{elementinfo6}).

\subsection{Organization}

The Organization module works in conjunction with the Role-Dependent Names and Name module. It is important to note that the MODS XML schema does not have an element named Organization.  In order to instill natural semantics into the ontology, we introduce the Organization module to replace the attribute ``Affiliation" and element ``Alternative Names". The concrete differences are outlined in Section \ref{sec:comparison}. Organization is used as the main entity which provides an Agent Role. Naturally, it makes sense for an organization to have a Name. In the case where an organization is referred to using different names, we denote the primary name with \textit{hasStandardizedName} and the rest of the names using \textit{hasName}.

\begin{figure}[h]
    \centering
    \includegraphics[width=\textwidth]{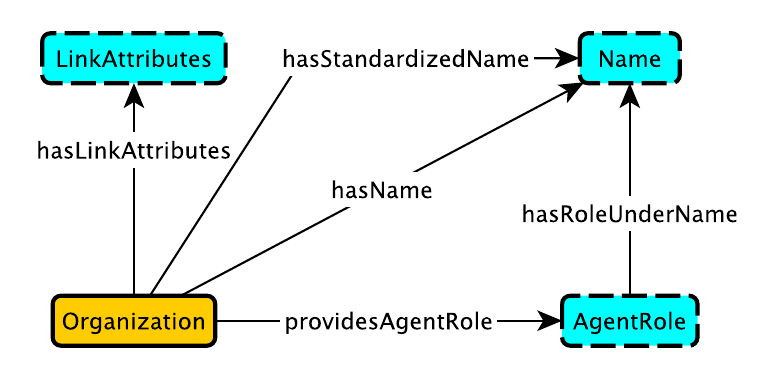}
    \caption{Schema Diagram for the Organization Module}
    \label{fig:organization}
\end{figure}

\subsubsection{Selected Axioms}

\begin{align}
  \textsf{Organization} &\sqsubseteq \mathord{\geq} 0\textsf{providesAgentRole.AgentRole} \label{organization1}\\
  \textsf{Organization} &\sqsubseteq \exists \textsf{hasName.Name} \label{organization2}\\
  \textsf{Organization} &\sqsubseteq \mathord{\geq} 0\textsf{hasStandardizedName.Name} \label{organization3}\\
  \top &\sqsubseteq \mathord{\leq} 1\textsf{hasLinkAttributes.} \top \label{organization4}\\
  \textsf{Organization} &\sqsubseteq \mathord{\geq} 0\textsf{hasLinkAttributes.LinkAttributes} \label{organization5}
\end{align}

It is not necessary that the Organization under description must provide an Agent Role. It can be referred in any general context, as such we say in (\ref{organization1}) that an Organization \emph{may} provide an Agent Role by using a \textit{structural tautology}. Furthermore, we argue that an Organization, naturally, must have a name and express that using an \textit{existential} in (\ref{organization2}). To distinguish between different names and the standardized name, we use (\ref{organization3}) to say that the Organization \emph{may} have a Standardized Name. Also an Organization \emph{may} have a set of Link Attributes to provide additional information (\ref{organization5}).


\subsection{Name}

The Name module is intended to be used for describing entities associated with the resource under description which may have one or more names. A necessary element of the Name module is Name Part. All the parts of a name (one or more) are described through Name Parts. In some cases, a name can refer to an acronym which is dictated by some Authority where the information regarding authority is expressed using Authority Information module. It is not uncommon for a name to have a specific form to display (e.g. Last name, First name), which is specified using Display Form. Furthermore, if a name has an associated identifier (e.g. ISBN, DOI), it is expressed using Name Identifier which is a sub-class of the module Identifier. 

\begin{sidewaysfigure}
    \includegraphics[width=.78\textwidth]{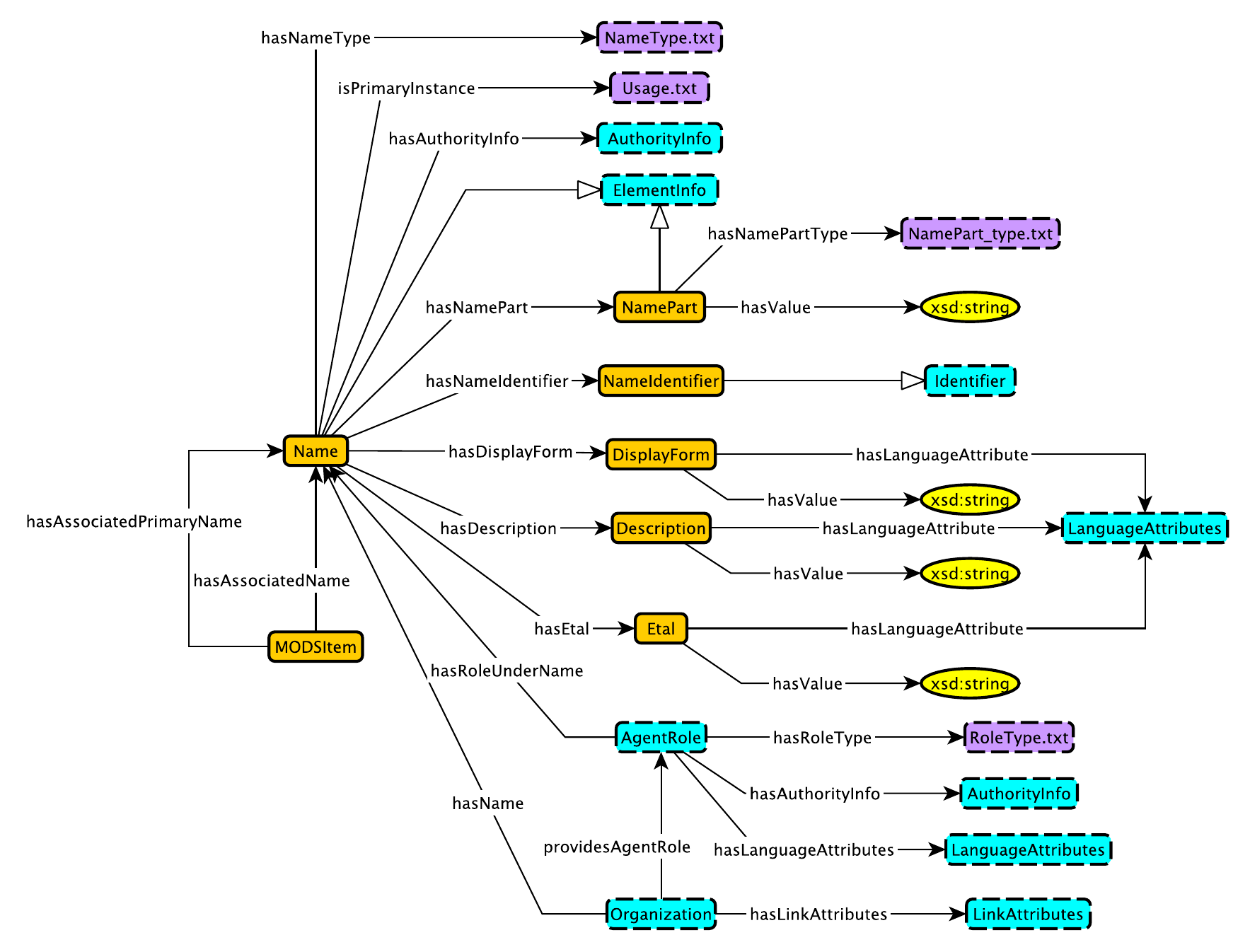}
    \caption{Schema Diagram for the Name Module}
    \label{fig:name}
\end{sidewaysfigure}

In the Name module, there are a few controlled vocabulary nodes (purple nodes in Figure \ref{fig:name}). To begin with, a Name can be assigned with a Name Type. MODS XML schema allows 4 name types: Personal, Corporate, Conference, Family. To let the user select a value from the available options, we make use of controlled vocabulary. Similarly, if among multiple instances of names, one particular name is to be regarded as the primary instance, the controlled vocabulary Usage is used to identify that. Another example of controlled vocabulary's usage can be seen in Name Part Type. To identify a part of name to be first name, middle name, or last name the Name Part Type controlled vocabulary can be used.

\subsubsection{Selected Axioms}

\begin{align}
  \textsf{Name} &\sqsubseteq \exists \textsf{hasNamePart.NamePart} \label{name1}\\
  \textsf{NamePart} &\sqsubseteq \textsf{hasNamePart}^-\textsf{.Name} \label{name2}\\
  \top &\sqsubseteq \mathord{\leq} 1\textsf{hasNamePart}^-\textsf{.} \top \label{name3}\\
  \textsf{Name} &\sqsubseteq \mathord{\geq} 0\textsf{hasNamePart.NamePart} \label{name4}\\
  \top &\sqsubseteq \forall \textsf{hasNamePartType.NamePartType.txt} \label{name5}\\
  \textsf{Name} &\sqsubseteq \mathord{\geq} 0\textsf{hasDescription.Description} \label{name6}\\
  \textsf{Name} &\sqsubseteq \mathord{\geq} 0\textsf{hasNameType.NameType.txt} \label{name7}\\
  \textsf{Name} &\sqsubseteq \mathord{\geq} 0\textsf{isPrimaryInstance.Usage.txt} \label{name8}\\
  \top &\sqsubseteq \mathord{\leq} 1\textsf{hasAuthorityInfo.} \top \label{name9}\\
  \textsf{Name} &\sqsubseteq \mathord{\geq} 0\textsf{hasAuthorityInfo.AuthorityInfo} \label{name10}\\
  \textsf{NamePart} &\sqsubseteq \textsf{ElementInfo} \label{name11}\\
  \textsf{NamePart} &\sqsubseteq \lnot(\exists \textsf{hasLinkAttributes.} \exists \textsf{hasID.} \top) \label{name12}\\
  \textsf{NameIdentifier} &\sqsubseteq \textsf{Identifier} \label{name13}
\end{align}

As described in the beginning of this module, a Name must have at least one NamePart. Otherwise, having a Name which does not have any string value as part of it would not be natural. We express this using an \textit{existential} in (\ref{name1}). On the other hand, to restrict the usage of NamePart outside of Name, we use an \textit{inverse existential} to convey that if there is a \textsf{hasNamePart} property, its domain must be a Name. A Name can also have any number of NameParts, to allow which we use \textit{structural tautology} in (\ref{name4}). Axioms (\ref{name1}) and (\ref{name4}) together mean that there can be one or more NameParts. 

The Name module is a \textit{sub-class} of Element Information (\ref{name11}) which says that a Name instance may have a set of Link Attributes and/or Language Attributes. One axiom to note here is (\ref{name12}) which essentially says that, an instance of a Name cannot have an ID which is a part of Link Attributes. The Link Attributes module has not been discussed here, we refer to the documentation for further details.

\subsection{Date Information and Date Attributes}

Date Information is a key module that has numerous usage within MMODS-O. A Bibliographic resource may have associated date information to express the timeline of creation, last updated, physical and/or digital origin information, etc. Throughout the MODS XML schema, all the date information under different names follow more or less a similar structure. That is why, we realized the necessity of having a Date Information module which conforms with our general intention of having a modular, reusable design. Primarily, a DateInfo instance may have a set of Language Attributes (e.g. date mentioned in multiple languages), some essential Date Attributes. We have created a Date Attributes module to further aid reusability and compact design. Another important aspect of the DateInfo module is that it must have a type of DateInfoType. Note, that there is no DateInfoType available in MODS XML schema. We outline the differences in detail in Section \ref{sec:comparison}.

\begin{figure}[h]
    \centering
    \includegraphics[width=\textwidth]{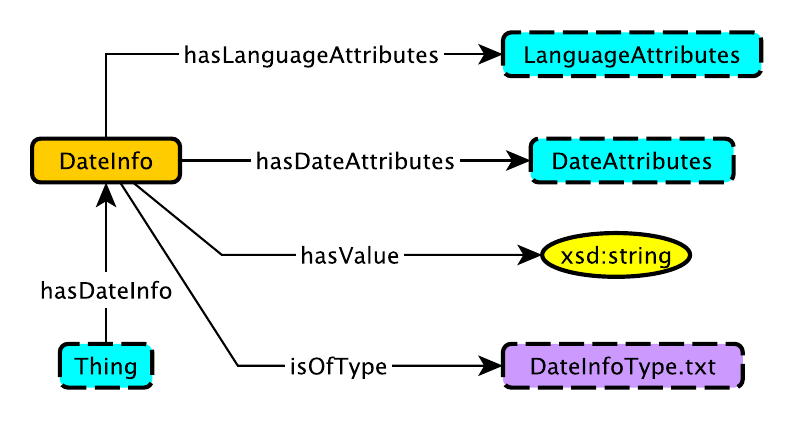}
    \caption{Schema Diagram for the Date Info Module}
    \label{fig:dateinfo}
\end{figure}

\begin{figure}[h]
    \centering
    \includegraphics[width=\textwidth]{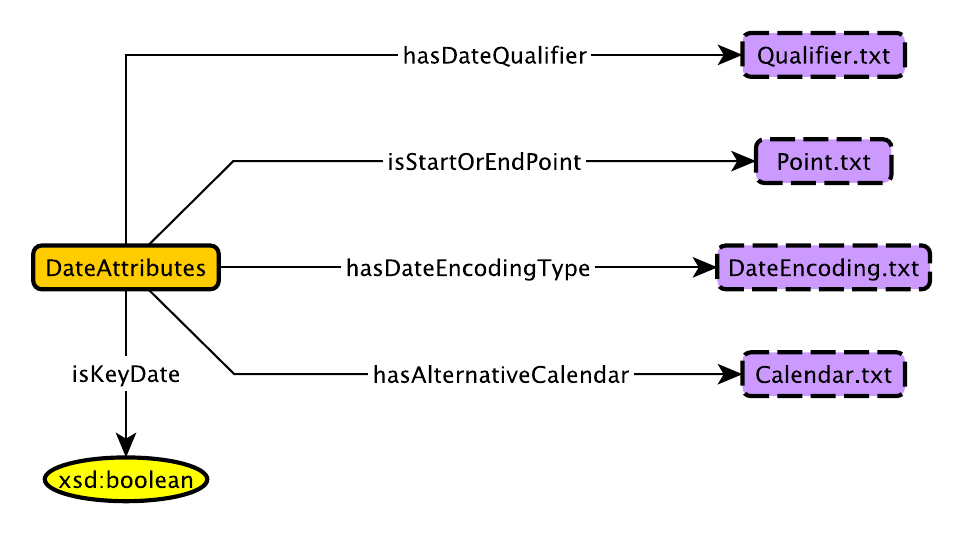}
    \caption{Schema Diagram for the Date Attributes Module}
    \label{fig:dateattributes}
\end{figure}

Different types of dates across the MODS XML schema generally offer a similar set of attributes, as such we make use of the DateAttributes module. The Qualifier identifies the date under description to be either \textit{approximate}, \textit{inferred}, or \textit{questionable} which is why this is a controlled vocabulary in Figure~\ref{fig:dateattributes}. The DateEncoding controlled vocabulary identifies the encoding type of the date (e.g. \textit{w3cdtf}, \textit{iso8601}). It is also possible to identify one DateInfo instance to be the Key Date among different instances of DateInfo using the DateAttributes with the property \textsf{isKeyDate} which provides a boolean value.

\subsubsection{Selected Axioms}

\begin{align}
  \top &\sqsubseteq \mathord{\leq} 1\textsf{hasDateInfo}^-\textsf{.} \top \label{dateinfo1}\\
  \textsf{Thing} &\sqsubseteq \mathord{\geq} 0\textsf{hasDateInfo.DateInfo} \label{dateinfo2}\\
  \textsf{DateInfo} &\sqsubseteq \exists \textsf{hasDateAttributes.DateAttributes} \label{dateinfo3}\\
  \top &\sqsubseteq \mathord{\leq} 1\textsf{hasDateAttributes.} \top \label{dateinfo4}\\
  \textsf{DateInfo} &\sqsubseteq \mathord{\geq} 0\textsf{hasDateAttributes.DateAttributes} \label{dateinfo5}\\
  \textsf{DateInfo} &\sqsubseteq \exists \textsf{isOfType.DateInfoType.txt} \label{dateinfo6}\\
  \textsf{DateInfo} &\sqsubseteq \exists \textsf{hasValue.xsd:string} \label{dateinfo7}\\
  \textsf{DateAttributes} &\sqsubseteq \mathord{\geq} 0\textsf{hasDateEncodingType.DateEncoding.txt} \label{dateinfo8}\\
  \textsf{DateAttributes} &\sqsubseteq \mathord{\geq} 0\textsf{isKeyDate.xsd:boolean} \label{dateinfo9}\\
  \textsf{DateAttributes} &\sqsubseteq \mathord{\geq} 0\textsf{isStartOrEndPoint.Point.txt} \label{dateinfo10}\\
  \textsf{DateAttributes} &\sqsubseteq \mathord{\geq} 0\textsf{hasAlternativeCalendar.Calendar.txt} \label{dateinfo11}
\end{align}

In order to formalize the intended use, the property \textsf{hasDateInfo} can only be associated with at most one instance of \emph{Thing}, expressed using an \textit{inverse functionality} (\ref{dateinfo1}) wherein a \emph{Thing} can have 0 or more instances of DateInfo, expressed using a \textit{structural tautology} (\ref{dateinfo2}). An instance of DateInfo must have exactly one set of DateAttributes which is conveyed by using a combination of \textit{existential} (\ref{dateinfo3}) and \textit{functionality} (\ref{dateinfo4}). Furthermore, a DateInfo must have a DateInfo type (\ref{dateinfo6}). The DateInfo type is a controlled vocabulary that contains a list of Date elements available in MODS XML schema, for example: \textsf{dateIssued}, \textsf{dateCreated}, \textsf{dateCaptured}, \textsf{dateModified}, \textsf{dateValid}, etc.

\bigskip

We have outlined 7 out of the 34 modules we have created as part of the MMODS-O ontology. In those 7 modules, we have only discussed the formal axioms which we considered the most interesting. The documentation contains a detailed description of all the modules including a comprehensive formalization.





\section{Related Work and Comparison with Previous Work}
\label{sec:comparison}

 To the best of our knowledge, there is very few published work available regarding ontologies based on MODS. 
The closest effort appears to be the MODS RDF Ontology\footnote{\url{https://www.loc.gov/standards/mods/modsrdf/primer.html}} available from Library of Congress pages. It appears to be a mostly straightfoward transcription of the XML schema without significant effort to make modifications to adjust to the ontology paradigm. We will use this for comparison; as it is very close to the MODS XML schema, we make only reference to the XML schema in the discussion.\footnote{We also found \url{http://arco.istc.cnr.it:8081/ontologies/MODS} which appears to be abandoned work-in progress without meaningful documentation.}
Our ontology design in many cases accounts for the natural relationships between entities which creates distinctions between our modeling and the MODS RDF Ontology and the XML schema.

 The Name entity in the XML schema raises a few issues when it comes to assessing the inherent meaning. For instance, the Name entity is treated to be both the name of a person and the person itself. There is no distinction between an individual and the individual having a name. This poses a lot of modeling issues and complications that can be overcome with an appropriate ontology-based approach. Questions arise such as: if an Agent is to be defined by its Name, what happens when the same Agent has multiple Names? Do we create separate instances of Name that in essence speak about the same Agent? How do we bind together the different names of the same Agent? In our case, we separate the notion of Agent and its Name which resolves the questions naturally. An Agent may have more than one name which is completely fine as is reflected in its axiomatization.

 Another issue we see with the Name entity is that, in XML schema a Name entity has an affiliation which is again another Name-like entity. Much like above, if we associate the name, the agent, and the affiliation all together with the name and agent, one may ask: if the agent has multiple names, do we create separate instances of names and write the same affiliations in all name instances? Perhaps more importantly, does it make more sense semantically to have an Organization entity that provides an affiliation? We argue that, an Agent, much less a Name, should not have an affiliation which is a Name, rather an Agent has an affiliation with an Organization, and that Organization will have a Name. 

 Furthermore, the XML schema states that the Name entity has a Role. We argue that it is more natural for an Agent to have a Name and for that same Agent to assume a particular Role. There are cases where it is possible for the same Agent to assume multiple roles under different pseudonyms. The XML schema and the existing RDF Ontology do not account for such intricate scenarios. The XML schema also allows for Names to have Alternative Names. It can be easily seen that it is not the Name which has Alternative Names, rather it is an Agent or an Organization which may have Alternative Names.

 Another instance where we argue that our approach is more modularized and has reusable aspects is concerning DateInfo. Both the XML schema and the MODS RDF Ontology use separate elements of dates to convey different use-cases of dates. Namely, dateIssued, dateCreated, dateCaptured, dateModified, dateValid, etc. What we have done instead is, we have created a common module for DateInfo, where for each of the use-cases of dates can just be defined as a type of date through the use of controlled vocabularies. This module also recognizes the fact that all date-related elements within MODS share the same set of attributes, which gives rise to the DateAttributes model.

 In our opinion, it is important to define and limit the applicability of modules within an ontology which we achieve through our carefully thought-out axiomatizations. It is imperative to leverage the different types of axioms available such as Scoped Domain, Scoped Range, Existential, Inverse Existential, Functionalities, Inverse Functionalities in order to formalize the scopes and boundaries. The existing RDF Ontology only uses Domain, Range, and Subproperties as formalization of the ontology, which in our opinion does often not suffice \cite{DBLP:books/ios/p/HitzlerK16}. 

\section{Conclusion}
\label{sec:conc}

We have presented the MMODS-O Ontology which has been developed from the MODS XML schema that has general use-cases in dealing with bibliographic metadata. We have developed the ontology in a way such that it is modularized, the distinct modules are reusable, and it paves the way for future improvement and module additions to the ontology. It incorporates modules that are concerned with Title information, Origin information, Geographic location, Target audience, Name, Subject, etc., of the resource under description. The ontology is serialized in OWL and has been formalized by extensive axiomatization. 

\medskip

\noindent\emph{Acknowledgement.} The authors acknowledge funding under the National Science Foundation grants 2119753 "RII Track-2 FEC: BioWRAP (Bioplastics With Regenerative Agricultural Properties): Spray-on bioplastics with growth synchronous decomposition and water, nutrient, and agrochemical management" and 2033521: "A1: KnowWhereGraph: Enriching and Linking Cross-Domain Knowledge Graphs using Spatially-Explicit AI Technologies."

\bibliographystyle{splncs04}
\bibliography{ref}

\begin{thebibliography}{10}
\providecommand{\url}[1]{\texttt{#1}}
\providecommand{\urlprefix}{URL }
\providecommand{\doi}[1]{https://doi.org/#1}

\bibitem{dlprimer}
Baader, F., Calvanese, D., Mcguinness, D., Nardi, D., Patel-Schneider, P.: The
  Description Logic Handbook: Theory, Implementation, and Applications (01
  2007)

\bibitem{rdf-tr}
Guha, R., Brickley, D.: {RDF} {S}chema {1.1}. {W3C} {R}ecommendation, W3C (Feb
  2014), https://www.w3.org/TR/2014/REC-rdf-schema-20140225/

\bibitem{Hitzler21}
Hitzler, P.: A review of the semantic web field. Commun. {ACM}  \textbf{64}(2),
   76--83 (2021)

\bibitem{DBLP:books/ios/p/HitzlerK16}
Hitzler, P., Krisnadhi, A.: On the roles of logical axiomatizations for
  ontologies. In: Hitzler, P., Gangemi, A., Janowicz, K., Krisnadhi, A.,
  Presutti, V. (eds.) Ontology Engineering with Ontology Design Patterns --
  Foundations and Applications, Studies on the Semantic Web, vol.~25, pp.
  73--80. {IOS} Press (2016). \doi{10.3233/978-1-61499-676-7-73}

\bibitem{Hitzler2010}
Hitzler, P., Kr{\"{o}}tzsch, M., Rudolph, S.: Foundations of Semantic Web
  Technologies. Chapman and Hall/CRC Press (2010)

\bibitem{xml-tr}
Maler, E., Bray, T., Paoli, J., Yergeau, F., Sperberg-McQueen, M.: Extensible
  {M}arkup {L}anguage ({XML}) 1.0 ({F}ifth {E}dition). {W3C} {R}ecommendation,
  W3C (Nov 2008), https://www.w3.org/TR/2008/REC-xml-20081126/

\bibitem{modsmetadata}
McCallum, S.H.: An introduction to the metadata object description schema
  {(MODS)}. In: Proceedings of the 10th Workshop on Ontology Design and
  Patterns {(WOP} 2019) co-located with 18th International Semantic Web
  Conference {(ISWC} 2019), Auckland, New Zealand, October 27, 2019. pp.
  82--88. Emerald Group Publishing Limited (2019),
  \url{https://doi.org/10.1108/07378830410524521}

\bibitem{owl-tr}
Parsia, B., Kr{\"{o}}tzsch, M., Hitzler, P., Rudolph, S., Patel-Schneider, P.:
  {OWL 2 Web Ontology Language Primer (Second Edition)}. {W3C}
  {R}ecommendation, W3C (Dec 2012),
  https://www.w3.org/TR/2012/REC-owl2-primer-20121211/

\bibitem{roledependentname}
Rayan, R., Shimizu, C., Hitzler, P.: An ontology design pattern for
  role-dependent names (2023), arXiv:2305.02077. Available from
  https://arxiv.org/abs/2305.02077

\bibitem{momo-swj}
Shimizu, C., Hammar, K., Hitzler, P.: Modular ontology modeling. Semantic Web
  \textbf{14}(3),  459--489 (2023)

\bibitem{modl}
Shimizu, C., Hirt, Q., Hitzler, P.: {MODL:} {A} modular ontology design
  library. In: Janowicz, K., Krisnadhi, A.A., Poveda{-}Villal{\'{o}}n, M.,
  Hammar, K., Shimizu, C. (eds.) Proceedings of the 10th Workshop on Ontology
  Design and Patterns {(WOP} 2019) co-located with 18th International Semantic
  Web Conference {(ISWC} 2019), Auckland, New Zealand, October 27, 2019. {CEUR}
  Workshop Proceedings, vol.~2459, pp. 47--58. CEUR-WS.org (2019),
  \url{https://ceur-ws.org/Vol-2459/paper4.pdf}

\end{thebibliography}

\end{document}